# CIB-SE-YOLOv8: Optimized YOLOv8 for Real-Time Safety Equipment Detection on Construction Sites

Xiaoyi Liu*
Ira A. Fulton Schools of Engineering
Arizona State University
Tempe, USA
*Corresponding author:
xliu472@asu.edu

Ruina Du
Ira A. Fulton Schools of Engineering
Arizona State University
Tempe, USA
ruinadu@asu.edu

Lianghao Tan
W. P. Carey School of Business
Arizona State University
Tempe, USA
ltan22@asu.edu

Junran Xu
Independent Researcher
Nanjing, China
xujunran86@gmail.com

Chen Chen
Department of Mechanical and Industrial Engineering
University of Illinois at Chicago
Chicago, USA
cchen215@uic.edu

Huangqi Jiang
Department of Psychological Sciences
Case Western Reserve University
Cleveland, USA
hxj244@case.edu

Saleh Aldwais
Department of Mechanical and Industrial Engineering
University of Illinois at Chicago
Chicago, USA
saldwa2@uic.edu

*Abstract*—Safety equipment is a crucial component in ensuring safety on construction sites, with helmets being particularly essential in reducing injuries and fatalities. Traditional methods, such as manual checks by superintendents and project managers, are inefficient and labor-intensive, often failing to prevent incidents related to the lack of safety gear. To address this, a new approach leveraging computer vision and deep learning has been developed, utilizing real-time detection techniques, specifically YOLO. This study utilizes the publicly available SHEL5K dataset for helmet detection tasks. The proposed CIB-SE-YOLOv8 model builds upon YOLOv8n by incorporating SE attention mechanisms and replacing certain C2f blocks with C2fCIB blocks. Compared to YOLOv8n, our model achieved a mAP50 of 88.4%, representing a 3.2% improvement. It also demonstrated a 0.5% increase in precision and a 3.9% increase in recall, significantly enhancing helmet detection performance. Additionally, the proposed model, with 2.68 million parameters and 7.6 GFLOPs, offers greater efficiency in real-time detection tasks compared to YOLOv8n's 3 million parameters and 8.1 GFLOPs, making it a more effective solution for improving construction site safety.

## I. INTRODUCTION

Construction safety is a critical concern in architecture and related industries. Ensuring that all construction workers wear proper safety equipment, such as helmets, is essential. However, traditional methods of manual monitoring by superintendents and project managers are inefficient. A new approach leveraging computer vision and deep learning technologies offers real-time detection capabilities. When a worker is not wearing a helmet, the camera equipped with computer vision technology can immediately detect this and issue an alert to the worker while reporting the incident to management. This system significantly reduces the risk of injury and other safety-related issues.

In this paper, we first compare the performance of YOLOv5n and YOLOv8n on helmet detection tasks on the publicly available dataset SHEL5k, which yields mAP50 scores of 84.7% and 85.2%, respectively. Based on the better performance and faster inference time of YOLOv8n, we further enhance the model by incorporating SE attention and replacing certain C2f modules with C2fCIB modules. The improved model, termed CIB-SE-YOLOv8, achieves a mAP50 of 88.4%, representing a notable improvement over both YOLOv8n and YOLOv5n. Moreover, this enhancement is achieved with fewer model parameters and shorter inference time, making it an excellent choice for real-time safety equipment detection on construction sites.

## II. RELATED WORK

### A. Machine Learning

Machine learning (ML) is heavily used nowadays in multiple industries. Some applications, such as data security [1], vehicle classification [2], recommendation systems [3], credit score prediction [4], and pose estimation of 3D objects [5], are great examples of the use of ML.

## B. Deep Learning

Advanced deep learning techniques have become essential in modern society, with applications including traffic flow prediction [6], image segmentation [7], pill identification [8], sentiment analysis [9], heterogeneous information network classification [10], and brain tumor segmentation [11].

## C. Safety Equipment for Construction Site

Safety equipment, such as helmets, plays a vital role in significantly reducing the risk of traumatic brain injuries (TBIs) and fatalities on construction sites. Wearing helmets greatly lowers the likelihood of injury, particularly in cases of work-related falls. Workers equipped with proper head protection are notably less susceptible to severe head injuries compared to those without helmets, highlighting the importance of helmet use in maintaining safety on construction sites [12].

## D. YOLO

*1) YOLOv5:* YOLOv5 (You Only Look Once, version 5) is a state-of-the-art object detection model released by the Ultralytics team. It includes several updates over previous YOLO architectures, featuring a modified CSPDarknet53 backbone with a Stem layer to reduce computational costs and an SPPF (Spatial Pyramid Pooling Fast) layer for efficient multi-scale feature extraction [13]. Additionally, it incorporates the AutoAnchor algorithm, which optimizes anchor boxes to enhance detection performance [13].

*2) YOLOv8:* YOLOv8, released in 2023 by Ultralytics, introduces several updates over previous versions. The anchor-free design and decoupled head enable better handling of complex scenes and improve performance on small or overlapping objects [13]. The C2f module replaces the CSPLayer, combining high-level features with contextual information to enhance detection accuracy [13]. The YOLO models can play a vital role in computer vision, both in real life and virtual reality (VR) [14] [15].

## III. METHODOLOGY

### A. Dataset

SHEL5K [16] is a publicly available dataset utilized in this study. The 5,000 images are divided into training, validation, and testing sets in a 7:2:1 ratio. Two classes are annotated: 15,051 instances of heads with helmets and 5,690 instances of heads without helmets. Inspired by Huang et al. [5], the SHEL5K dataset was selected for its incorporation of diverse background images, which simulate various environments, thereby enhancing the model's adaptability to real-world scenarios.

### B. Image Preprocessing

The images will be scaled down to a target size of 416x416. If the aspect ratio is not 1:1, empty areas will be filled by mirroring the image's edges and duplicating nearby pixels to create a seamless look. This process avoids empty or padded sections and prevents distortion of the images. More examples of images can be found in Figure 4.

## C. Learning Rate and Epoch

All research in this paper employs the Stochastic Gradient Descent (SGD) optimizer with a learning rate of 0.01 and a momentum of 0.9. The learning rate controls the adjustment size for each step, while momentum provides smoother updates for quicker optimization.

Following the approach of Huang et al. [5], a patience mechanism was implemented for early stopping if no improvements in validation performance were observed over 10 consecutive epochs. The best-performing model is tracked throughout the training process. This approach helps to avoid unnecessary computations while preventing both underfitting and overfitting.

TABLE I. PERFORMANCE METRICS FOR YOLO ON HELMET DETECTION

| Model | Precision | Recall | mAP50 | mAP 50-95 |
|---|---|---|---|---|
| YOLOv5n | 0.89 | 0.779 | 0.847 | 0.505 |
| YOLOv8n | 0.889 | 0.774 | 0.852 | 0.501 |
| **CIB-SE-YOLOv8(ours)** | **0.894** | **0.813** | **0.884** | **0.54** |

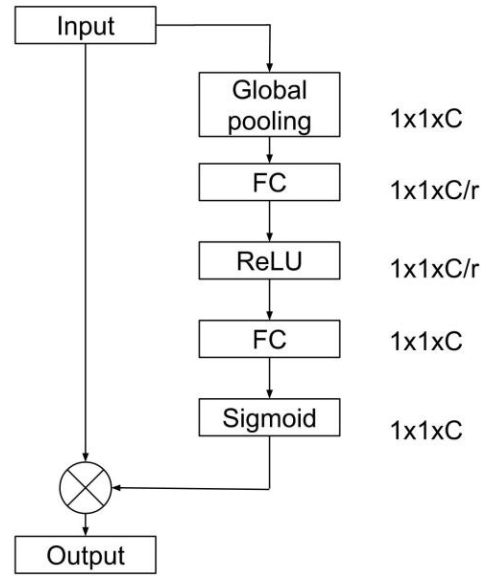

Figure 1. SE Attention

## D. YOLOv5n vs YOLOv8n

YOLOv5n and YOLOv8n reached mAP50 scores of 84.7% and 85.2%, respectively, with YOLOv8n showing a slight improvement in mAP50. As shown in Table I, YOLOv5n exhibits marginally higher precision and recall compared to YOLOv8n. Despite the overall performance of both models being comparable, YOLOv8n offers faster inference speed.

Therefore, YOLOv8n will be used as the baseline model in the following study.

*E. SE Attention Mechanism*

Squeeze-and-excitation (SE) attention is a neural network module introduced to improve the performance of convolutional neural networks (CNNs). It enables the model to focus more on important feature maps, with a relatively small increase in the number of parameters. SE attention consists of two main steps: squeeze and excitation. During the squeezing step, global average pooling is applied across all feature maps, producing a channel-wise descriptor [17]. In the excitation step, this descriptor is passed through two fully connected layers with non-linear activations, followed by a sigmoid function [17]. The resulting scale is then applied to the original feature maps, enhancing the CNN's ability to focus on the most relevant features [17]. More details about the architecture of SE attention can be found in Figure 1.

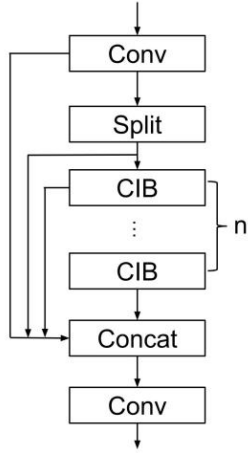

Figure 2. C2fCIB

*F. C2fCIB module*

C2fCIB replaces the bottleneck modules in C2f with Compact Inverted Block (CIB) modules while maintaining the overall structure. CIB is designed to be a highly efficient building block for deep learning models [18]. Additionally, a rank-guided strategy is employed to minimize computational costs without sacrificing performance [18]. Further details about the architecture of C2fCIB can be found in Figure 2.

*G. Proposed Deep Learning Models*

For our proposed model, CIB-SE-YOLOv8, we use YOLOv8n as the baseline model. In the backbone, we replaced the C2f modules at layers 6 and 8 with C2fCIB modules. This enhances the model's ability to understand the context and capture spatial relationships between features, helping the network focus more precisely on small objects, such as helmets in our case. Additionally, SE attention layers were added after the C2f blocks at layers 15, 18, and 21. By incorporating SE layers at the small, medium, and large feature map stages, the SE attention helps the model identify which features are most important before passing them to the detection layers. More details can be found in Figure 3.

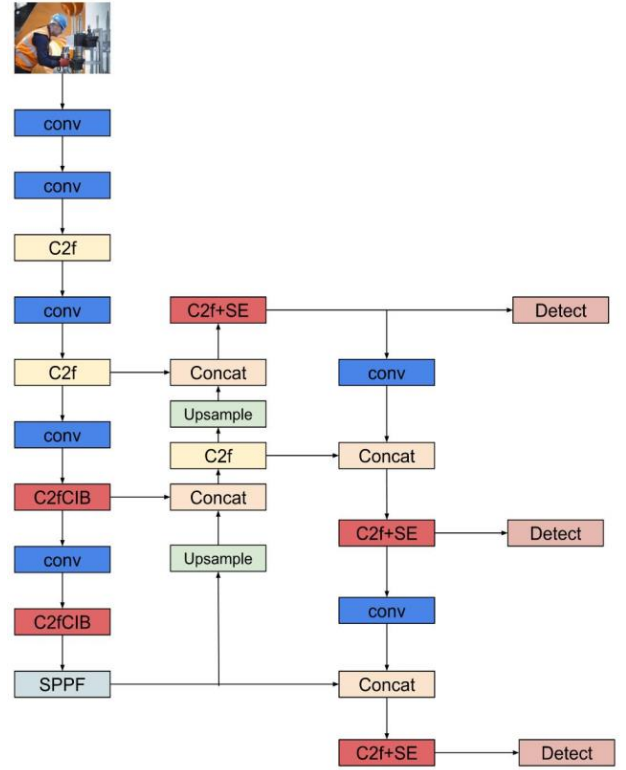

Figure 3. CIB-SE-YOLOv8

TABLE II. ABLATION STUDY

| Model | Precision | Recall | mAP50 | mAP 50-95 |
|---|---|---|---|---|
| YOLOv8n | 0.889 | 0.774 | 0.852 | 0.501 |
| YOLOv8n-SE | **0.909** | 0.799 | 0.878 | 0.539 |
| YOLOv8n-C2fCIB | 0.889 | 0.807 | 0.876 | 0.533 |
| **CIB-SE-YOLOv8(ours)** | 0.894 | **0.813** | **0.884** | **0.54** |

*H. Ablation Study*

By incorporating the SE attention mechanism and replacing C2f blocks with C2fCIB blocks, the model's mAP50 improved from 85.2% to 87.8% and 87.6%, respectively. These enhancements demonstrate a positive impact on the model's performance. Further development by combining both changes led to an additional improvement, raising the mAP50 to 88.4%. Detailed results are provided in Table II.

*I. Performance Measurements*

Mean Average Precision (mAP), Precision, and Recall are the primary metrics used to measure performance in this study, while mAP50-95 serves as a supplementary metric. The equations for Precision and Recall are provided below, where TP represents True Positives, FP represents False Positives, and FN represents False Negatives.

$$Precision = \frac{TP}{TP + FP} \quad (1)$$

$$Recall = \frac{TP}{TP + FN} \quad (2)$$

mAP is an essential metric for object detection tasks, measuring a model's effectiveness across all classes. It calculates the area under the precision-recall curve for each class, with a higher mAP indicating a better balance between precision and recall, reflecting the model's effectiveness in accurately detecting and classifying objects. mAP50 measures the mean average precision at an Intersection over Union (IoU) threshold of 0.50, meaning that a predicted bounding box is considered correct if it overlaps with the ground truth box by at least 0.5. For mAP, the corresponding equation is shown below, where $N$ is the number of classes and $AP_i$ represents the average precision for each class $i$. AP evaluates the area under the precision-recall curve, indicating a model's performance for a specific class.

$$mAP = \frac{1}{N} \sum_{i=1}^{N} AP_i \quad (3)$$

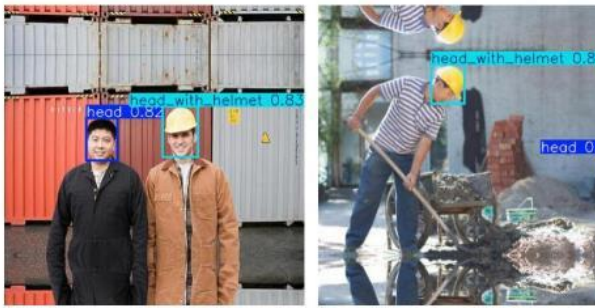
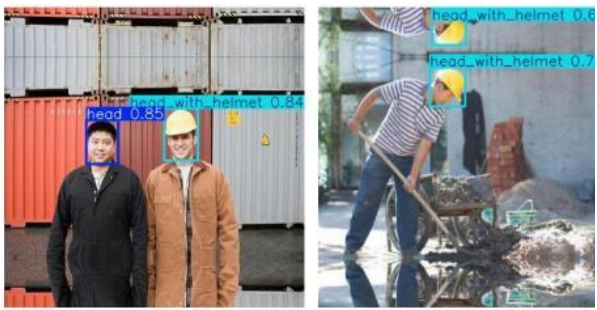

Figure 4. YOLOv8n vs CIB-SE-YOLOv8 (Ours)

## IV. EVALUATION AND DISCUSSION OF RESULTS

CIB-SE-YOLOv8, when compared to the baseline YOLOv8 model, demonstrated improvements in precision by 0.5%, recall by 3.9%, mAP50 by 3.2%, and mAP50-95 by 3.9%. Additionally, our proposed model has a reduced parameter count, with 2,683,222 parameters and 7.6 GFLOPs, compared to YOLOv8's 3,006,038 parameters and 8.1 GFLOPs.

YOLOv8n:

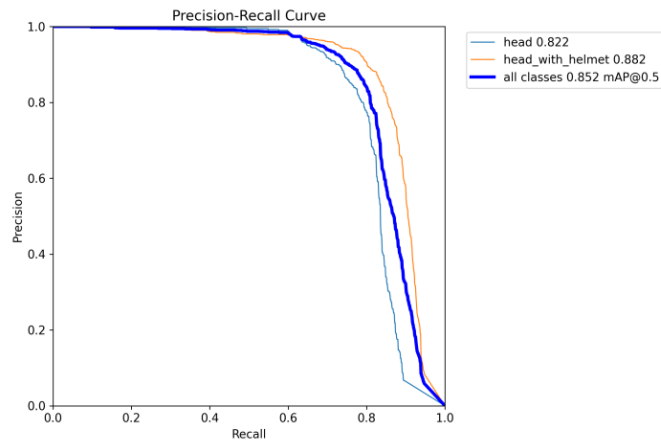

CIB-SE-YOLOv8:

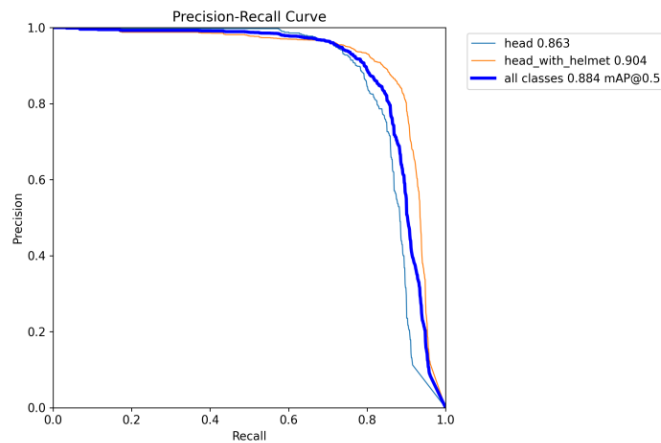

Figure 5. PR_Curve: YOLOv8n vs CIB-SE-YOLOv8 (Ours)

These results indicate that CIB-SE-YOLOv8 not only offers enhanced performance but also achieves greater efficiency in terms of parameters and computational load, making it well-suited for helmet detection tasks. From Figure 4, it is evident that our proposed model achieved a higher prediction accuracy on the left image. In the right image, our model correctly identifies both helmets without mistakenly classifying the background as a head. From Figure 5, a comparison of the Precision-Recall curve further demonstrates our model's improvement. Moreover, YOLOv8n is capable of achieving real-time detection on most GPUs, maintaining performance of at least 30 frames per second (FPS). On CPUs, a frame-dropping technique can be employed to ensure real-time detection by selectively processing fewer frames. Additionally, CIB-SE-YOLOv8, with its reduced parameter count and lower GFLOPs, offers even greater efficiency for real-time detection tasks, making it particularly suitable for our construction site helmet detection scenario.

## V. CONCLUSION

CIB-SE-YOLOv8 shows significant improvements over YOLOv8n across various metrics, including precision, recall, and mAP, while also having a smaller parameter size and achieving faster GFLOPs. These improvements greatly enhance its efficiency for detecting helmets in real-time on construction sites. The model enables 24/7 monitoring, significantly easing the workload for superintendents or project managers. Moreover, it helps reduce the risk of injuries or fatalities by promptly identifying and warning individuals who are not wearing protective equipment like helmets, thus enhancing overall site safety.

## REFERENCES

[1] Y. Weng and J. Wu, "Leveraging artificial intelligence to enhance data security and combat cyber attacks," *Journal of Artificial Intelligence General science (JAIGS) ISSN: 3006-4023*, vol. 5, no. 1, pp. 392–399, 2024.

[2] X. Li, J. Chang, T. Li, W. Fan, Y. Ma, and H. Ni, "A vehicle classification method based on machine learning," 2024.

[3] Z. Wu, X. Wang, S. Huang, H. Yang, D. Ma *et al.*, "Research on prediction recommendation system based on improved markov model," *Advances in Computer, Signals and Systems*, vol. 8, no. 5, pp. 87–97, 2024.

[4] Q. Xing, C. Yu, S. Huang, Q. Zheng, X. Mu, and M. Sun, "Enhanced credit score prediction using ensemble deep learning model," *arXiv preprint arXiv:2410.00256*, 2024.

[5] S. Huang, Y. Song, Y. Kang, and C. Yu, "Ar overlay: Training image pose estimation on curved surface in a synthetic way," *arXiv preprint arXiv:2409.14577*, 2024.

[6] Z. Wu, "Deep learning with improved metaheuristic optimization for traffic flow prediction," *Journal of Computer Science and Technology Studies*, vol. 6, no. 4, pp. 47–53, 2024.

[7] Q. Yang, Z. Wang, S. Liu, and Z. Li, "Research on improved u-net based remote sensing image segmentation algorithm," *arXiv preprint arXiv:2408.12672*, 2024.

[8] B. Dang, W. Zhao, Y. Li, D. Ma, Q. Yu, and E. Y. Zhu, "Real-time pill identification for the visually impaired using deep learning," in *2024 6th International Conference on Communications, Information System and Computer Engineering (CISCE)*, 2024, pp. 552–555.

[9] W. Zhu and T. Hu, "Twitter sentiment analysis of covid vaccines," in *2021 5th International Conference on Artificial Intelligence and Virtual Reality (AIVR)*, 2021, pp. 118–122.

[10] Z. Wu, "Mpgaan: Effective and efficient heterogeneous information network classification," *Journal of Computer Science and Technology Studies*, vol. 6, no. 4, pp. 08–16, 2024.

[11] Q. Zhang, W. Qi, H. Zheng, and X. Shen, "Cu-net: a u-net architecture for efficient brain-tumor segmentation on brats 2019 dataset," arXiv preprint arXiv:2406.13113, 2024.

[12] S. C. Kim, Y. S. Ro, S. D. Shin, and J. Y. Kim, "Preventive effects of safety helmets on traumatic brain injury after work-related falls," *International journal of environmental research and public health*, vol. 13, no. 11, p. 1063, 2016.

[13] J. Terven, D.-M. Cordova-Esparza, and J.-A. Romero-González, "A comprehensive review of yolo architectures in computer vision: From yolov1 to yolov8 and yolo-nas," *Machine Learning and Knowledge Extraction*, vol. 5, no. 4, pp. 1680–1716, 2023.

[14] X. Yang, Y. Kang, and X. Yang, "Retargeting destinations of passive props for enhancing haptic feedback in virtual reality," in *2022 IEEE Conference on Virtual Reality and 3D User Interfaces Abstracts and Workshops (VRW)*. IEEE, 2022, pp. 618–619.

[15] Y. Kang, Y. Xu, C. Chen, G. Li, and Z. Cheng, "6: Simultaneous tracking, tagging and mapping for augmented reality," in SID Symposium Digest of Technical Papers, vol. 52. Wiley Online Library, 2021, pp. 31–33.

[16] Database, "Shel5k new dataset," https://universe.roboflow.com/database-sjrvw/shel5k-new, 2022

[17] J. Hu, L. Shen, and G. Sun, "Squeeze-and-excitation networks," in Proceedings of the IEEE conference on computer vision and pattern recognition, 2018, pp. 7132–7141.

[18] A. Wang *et al.*, "Yolov10: Real-time end-to-end object detection," *arXiv preprint*, arXiv:2405.14458, 2024.